\documentclass{article}

\usepackage{microtype}
\usepackage{graphicx}
\usepackage{booktabs} %

\usepackage{hyperref}

\usepackage[preprint]{icml2024}

\usepackage{amsmath}
\usepackage[frozencache,cachedir=minted-cache]{minted}
\usepackage{amssymb}
\usepackage[T1]{fontenc}
\usepackage{inconsolata}
\usepackage{framed}
\usepackage{mathtools}
\usepackage[export]{adjustbox}
\usepackage{float}
\usepackage{placeins}
\usepackage{afterpage}
\usepackage{amsthm,multirow}
\usepackage{xcolor}         %
\usepackage{xspace}
\usepackage{subcaption}

\usepackage[capitalize,noabbrev]{cleveref}

\theoremstyle{plain}

\theoremstyle{definition}

\theoremstyle{remark}

\newcommand{\rulesep}{\unskip\ \vrule width 0.8pt\ }
\usepackage[textsize=tiny]{todonotes}

\newsavebox{\mintedbox}
\begin{document}

\twocolumn[
\icmltitle{Disentangled 3D Scene Generation with Layout Learning}

\icmlsetsymbol{equal}{*}
\begin{icmlauthorlist}
\icmlauthor{Dave Epstein}{berk,goog}
\icmlauthor{Ben Poole}{goog}
\icmlauthor{Ben Mildenhall}{goog}
\icmlauthor{Alexei A. Efros}{berk}
\icmlauthor{Aleksander Holynski}{berk,goog}
\end{icmlauthorlist}
\icmlaffiliation{berk}{Department of Computer Science, UC Berkeley}
\icmlaffiliation{goog}{Google Research}

\icmlcorrespondingauthor{}{dave@eecs.berkeley.edu}

\icmlkeywords{text-to-3d, disentanglement, unsupervised learning, object discovery}

\vskip 0.3in
]

\printAffiliationsAndNotice{}  %

\begin{abstract}

We introduce a method to generate 3D scenes that are disentangled into their component objects. This disentanglement is unsupervised, relying only on the knowledge of a large pretrained text-to-image model. Our key insight is that objects can be discovered by finding parts of a 3D scene that, when rearranged spatially, still produce valid configurations of the same scene. 
Concretely, our method jointly optimizes multiple NeRFs from scratch---each representing its own object---along with a \emph{set of layouts} that composite these objects into scenes. We then encourage these composited scenes to be in-distribution according to the image generator. We show that despite its simplicity, our approach successfully generates 3D scenes decomposed into individual objects, enabling new capabilities in text-to-3D content creation. See our project page for results and an interactive demo: \url{https://dave.ml/layoutlearning/}

\end{abstract}

\begin{figure*}[t!]
    \input{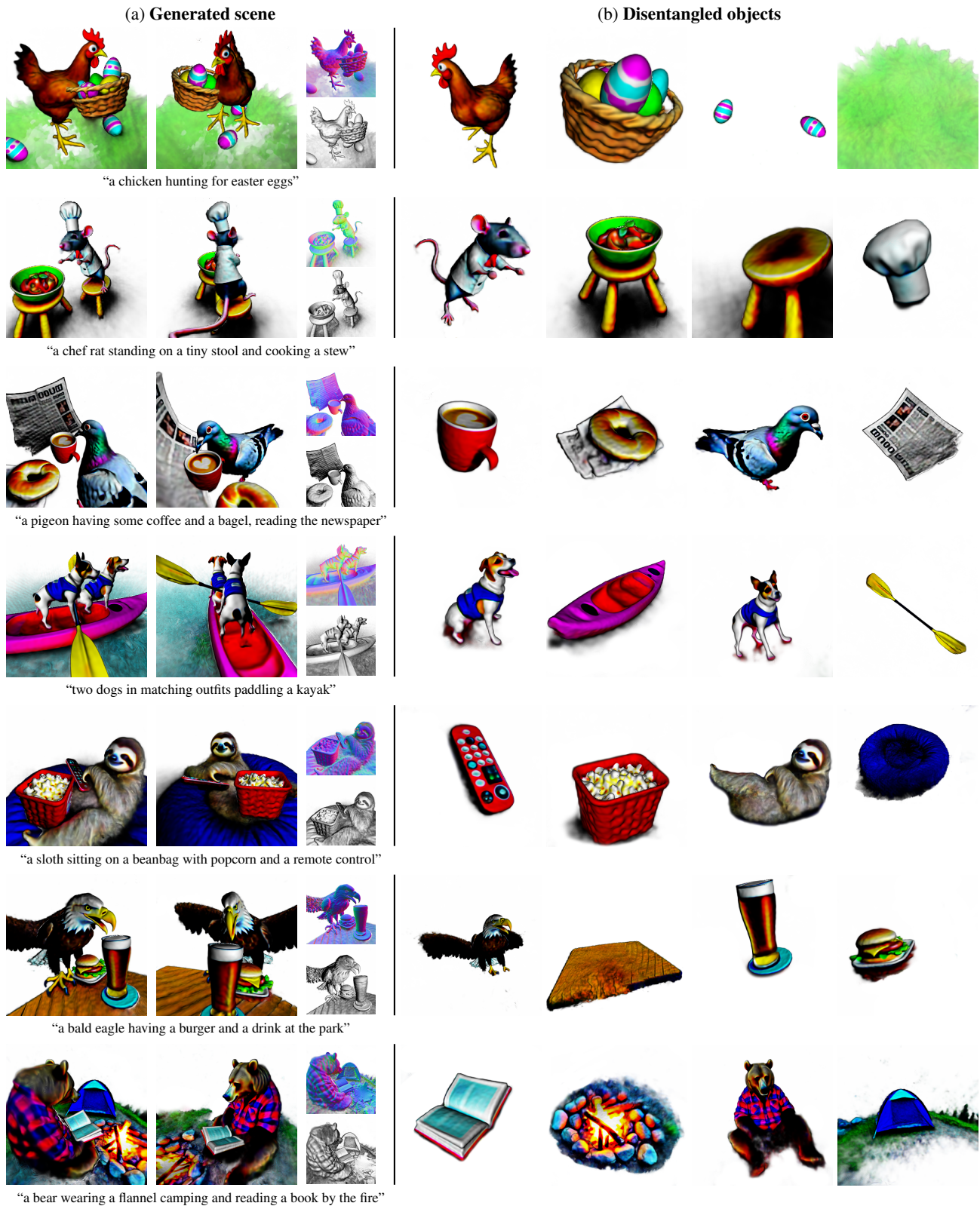}%
    \caption{\textbf{Layout learning generates disentangled 3D scenes} given a text prompt and a pretrained text-to-image diffusion model. We learn an entire 3D scene (left, shown from two views along with surface normals and a textureless render) that is composed of multiple NeRFs (right) representing different objects and arranged according to a learned layout.
    }

    \label{fig:main_results}
     \vspace{-11pt}
\end{figure*}
\section{Introduction}
A remarkable ability of many seeing organisms is object individuation \cite{piaget1952origins}, the ability to discern separate objects from light projected onto the retina~\cite{wertheimer1938laws}. Indeed, from a very young age, humans and other creatures are able to organize the physical world they perceive into the three-dimensional entities that comprise it \cite{spelke1990principles, wilcox1999object, crowscanseeobjects}.
The analogous task of object discovery has captured the attention of the artificial intelligence community from its very inception \cite{roberts1963machine, ohta1978analysis}, since agents that can autonomously parse 3D scenes into their component objects are better able to navigate and interact with their surroundings.

Fifty years later, generative models of images are advancing at a frenzied pace \cite{nichol2021glide,ramesh2022hierarchical,saharia2022photorealistic,yu2022scaling,chang2023muse}. While these models can generate high-quality samples, their internal workings are hard to interpret, and they do not explicitly represent the distinct 3D entities that make up the images they create. Nevertheless, the priors learned by these models have proven incredibly useful across various tasks involving 3D reasoning \cite{hedlin2023unsupervised,ke2023repurposing,liu2023zero,luo2023diffusion, wu2023reconfusion}, suggesting that they may indeed be capable of decomposing generated content into the underlying 3D objects depicted.

One particularly exciting application of these text-to-image networks is 3D generation, leveraging the rich distribution learned by a diffusion model to optimize a 3D representation, {\em e.g.} a neural radiance field (NeRF, \citeauthor{mildenhall2020nerf}, \citeyear{mildenhall2020nerf}), such that rendered views resemble samples from the prior.
This technique allows for text-to-3D generation without any 3D supervision \cite{sds, sjc}, but most results focus on simple prompts depicting just one or two isolated objects \cite{lin2023magic3d, wang2023prolificdreamer}. 

Our method builds on this work to generate complex scenes that are automatically disentangled into the objects they contain. To do so, we instantiate and render {\em multiple NeRFs} for a given scene instead of just one, encouraging the model to use each NeRF to represent a separate 3D entity.
At the crux of our approach is an intuitive definition of objects as parts of a scene that can be manipulated independently of others while keeping the scene ``well-formed'' \cite{biederman1981semantics}. We implement this by learning a set of different layouts---3D affine transformations of every NeRF---which must yield composited scenes that render into in-distribution 2D images given a text prompt \cite{sds}.

We find that this lightweight inductive bias, which we term {\em layout learning}, results in surprisingly effective object disentanglement in generated 3D scenes (Figure \ref{fig:main_results}), enabling object-level scene manipulation in the text-to-3D pipeline. We demonstrate the utility of layout learning on several tasks, such as building a scene around a 3D asset of interest, sampling different plausible arrangements for a given set of assets, and even parsing a provided NeRF into the objects it contains, all without any supervision beyond just a text prompt.
We further quantitatively verify that, despite requiring no auxiliary models or per-example human annotation, the object-level decomposition that emerges through layout learning is meaningful and outperforms baselines.  
\clearpage
\begin{figure*}[t!]
     \centering
    \includegraphics[width=0.9\textwidth]{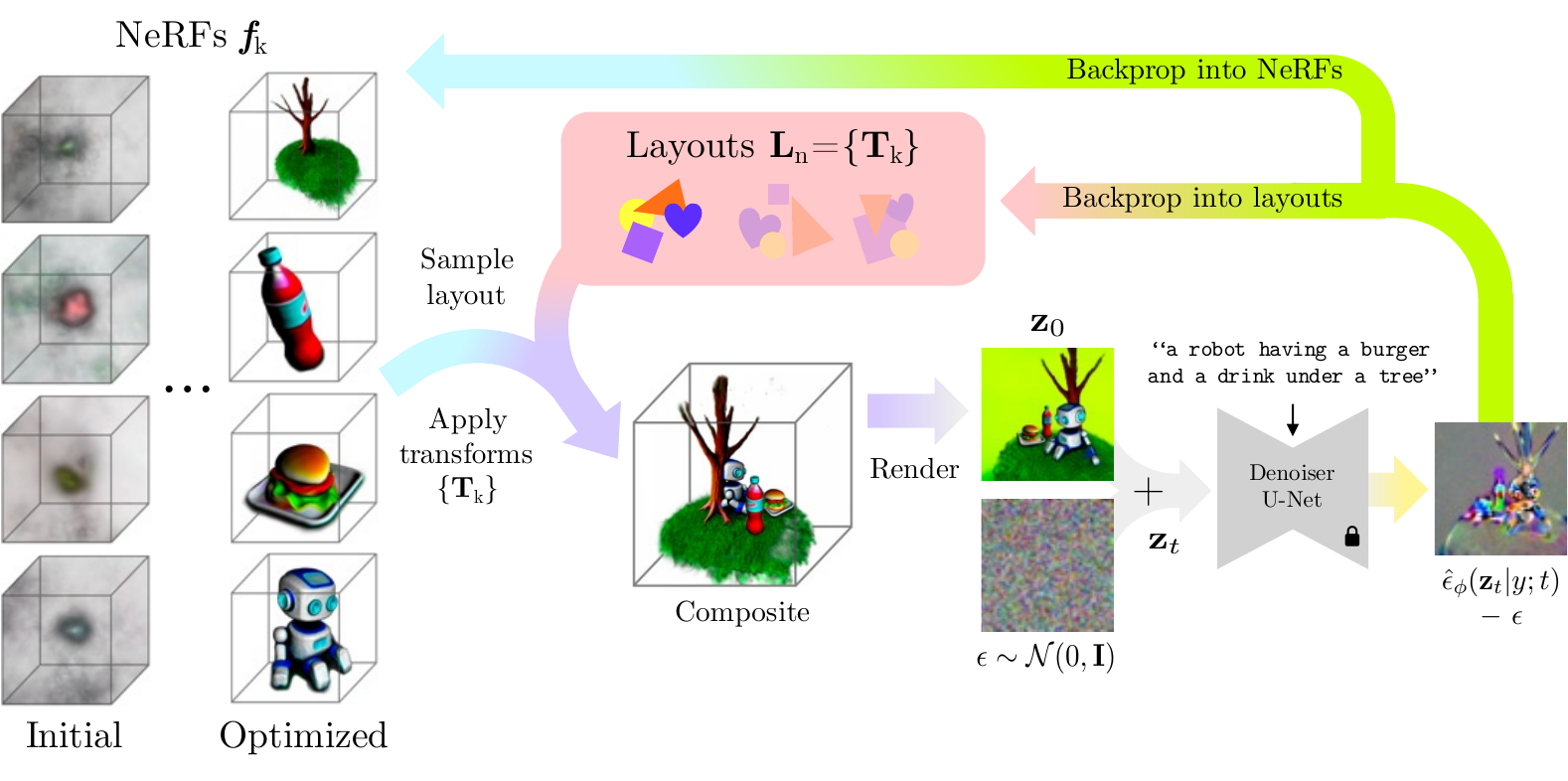}
    \vspace{-0.5em}
    \caption{\textbf{Method.} Layout learning works by optimizing $K$ NeRFs $f_k$ and learning $N$ different layouts $\mathbf{L}_n$ for them, each consisting of per-NeRF affine transforms $\mathbf{T}_k$. Every iteration, a random layout is sampled and used to transform all NeRFs into a shared coordinate space. The resultant volume is rendered and optimized with score distillation sampling \cite{sds} as well as per-NeRF regularizations to prevent degenerate decompositions and geometries \cite{Barron_2022_CVPR}. This simple structure causes object disentanglement to emerge in generated 3D scenes.}
    \label{fig:method}
     \vspace{-0.75em}
\end{figure*}
Our key contributions are as follows:
\begin{itemize}
\itemsep 0.15em 
    \item We introduce a simple, tractable definition of objects as portions of a scene that can be manipulated independently of each other and still produce valid scenes.
    \item We incorporate this notion into the architecture of a neural network, enabling the compositional generation of 3D scenes by optimizing a set of NeRFs as well as a set of layouts for these NeRFs.
    \item We apply layout learning to a range of novel 3D scene generation and editing tasks, demonstrating its ability to disentangle complex data despite requiring no object labels, bounding boxes, fine-tuning, external models, or any other form of additional supervision.
\end{itemize}

\section{Background}
\subsection{Neural 3D representations}
To output three-dimensional scenes, we must use an architecture capable of modeling 3D data, such as a neural radiance field (NeRF, \citeauthor{mildenhall2020nerf}, \citeyear{mildenhall2020nerf}). We build on MLP-based NeRFs \cite{Barron_2021_ICCV}, that represent a volume using an MLP $f$ that maps from a point in 3D space $\boldsymbol{\mu}$ to a density $\tau$ and albedo $\boldsymbol{\rho}$: $$(\tau , \boldsymbol{\rho}) = f(\boldsymbol{\mu}; \theta).$$
We can differentiably render this volume by casting a ray $\mathbf{r}$ into the scene, and then alpha-compositing the densities and colors at sampled points along the ray to produce a color and accumulated alpha value. For 3D reconstruction, we would optimize the colors for the rendered rays to match a known pixel value at an observed image and camera pose, but for 3D generation we sample a random camera pose, render the corresponding rays, and score the resulting image using a generative model.

\subsection{Text-to-3D using 2D diffusion models}
Our work builds on text-to-3D generation using 2D diffusion priors \citep{sds}. These methods turn a diffusion model into a loss function that can be used to optimize the parameters of a 3D representation. Given an initially random set of parameters $\theta$, at each iteration we randomly sample a camera $c$ and render the 3D model to get an image $x = g(\theta, c)$. We can then score the quality of this rendered image given some conditioning text $y$ by evaluating the score function of a noised version of the image $z_t=\alpha_t x + \sigma_t \epsilon$ using the pretrained diffusion model $\hat\epsilon(z_t; y, t)$.  We update the parameters of the 3D representation using score distillation:
\begin{equation}
   \nabla_\theta \mathcal{L}_\text{SDS}(\theta) = \mathbb{E}_{t,\epsilon,c}\left[w(t)(\hat\epsilon(z_t; y, t) - \epsilon)\frac{\partial x}{\partial \theta}\right]
\end{equation}
where $w(t)$ is a noise-level dependent weighting. 

SDS and related methods enable the use of rich 2D priors obtained from large text-image datasets to inform the structure of 3D representations. However, they often require careful tuning of initialization and hyperparameters to yield high quality 3D models, and past work has optimized these towards object generation. The NeRF is initialized with a Gaussian blob of density at the origin, biasing the optimization process to favor an object at the center instead of placing density in a skybox-like environment in the periphery of the 3D representation. Additionally, bounding spheres are used to prevent creation of density in the background. The resulting 3D models can produce high-quality individual objects, but often fail to generate interesting scenes, and the resulting 3D models are a single representation that cannot be easily split apart into constituent entities.

\section{Method}
\label{sec:method}
To bridge the gap from monolithic 3D representations to scenes with multiple objects, we introduce a more expressive 3D representation. Here, we learn multiple NeRFs along with a set of layouts, {\em i.e.} valid ways to arrange these NeRFs in 3D space. We transform the NeRFs according to these layouts and composite them, training them to form high-quality scenes as evaluated by the SDS loss with a text-to-image prior. This structure causes each individual NeRF to represent a different object while ensuring that the composite NeRF represents a high-quality scene. See Figure~\ref{fig:method} for an overview of our approach.

\subsection{Compositing multiple volumes}
We begin by considering perhaps the most na\"ive approach to generating 3D scenes disentangled into separate entities. We simply declare $K$ NeRFs $\{f_k\}$---each one intended to house its own object---and jointly accumulate densities from all NeRFs along a ray, proceeding with training as normal by rendering the composite volume. This can be seen as an analogy to set-latent representations \cite{locatello2020object, jaegle2021perceiverio, jaegle2021perceiver, jabri2023scalable}, which have been widely explored in other contexts. In this case, rather than arriving at the final albedo $\boldsymbol{\rho}$ and density $\tau$ of a point $\boldsymbol{\mu}$ by querying one 3D representation, we query $K$ such representations, obtaining a set $\{\boldsymbol{\rho}_k, \tau_k\}_{k=1}^K$. The final density at $\boldsymbol{\mu}$ is then $\tau^\prime = \sum \tau_k$ and the final albedo is the density-weighted average $\boldsymbol{\rho}^\prime = \sum \frac{\tau_k}{\tau^\prime} \boldsymbol{\rho}_k$. 

This formulation provides several potential benefits. First, it may be easier to optimize this representation to generate a larger set of objects, since there are $K$ distinct 3D Gaussian density spheres to deform at initialization, not just one. Second, many representations implicitly contain a local smoothness bias \cite{tancik2020fourier} which is helpful for generating objects but not spatially discontinuous scenes. Thus, our representation might be inclined toward allocating each representation toward a spatially smooth entity, {\em i.e.} an object. %

However, just as unregularized sets of latents are often highly uninterpretable, simply spawning $K$ instances of a NeRF does not produce meaningful decompositions. In practice, we find each NeRF often represents a random point-cloud-like subset of 3D space (Fig. \ref{fig:ablation}).

To produce scenes with disentangled objects, we need a method to encourage each 3D instance to represent a coherent object, not just a different part of 3D space.

\subsection{Layout learning}
We are inspired by other unsupervised definitions of objects that operate by imposing a simple inductive bias or regularization in the structure of a model's latent space, {\em e.g.} query-axis softmax attention \cite{locatello2020object}, spatial ellipsoid feature maps \cite{epstein2022blobgan}, and diagonal Hessian matrices \cite{peebles2020hessian}. In particular, \citet{niemeyer2021giraffe} learn a 3D-aware GAN that composites multiple NeRF volumes in the forward pass, where the latent code contains a random affine transform for each NeRF's output. Through this structure, each NeRF learns to associate itself with a different object, facilitating the kind of disentanglement we are after. However, their approach relies on pre-specified independent distributions of each object's location, pose, and size, preventing scaling beyond narrow datasets of images with one or two objects and minimal variation in layout. %

In our setting, not only does the desired output comprise numerous open-vocabulary, arbitrary objects, but these objects {\em must be arranged in a particular way} for the resultant scene to be valid or ``well-formed'' \cite{biederman1982scene}. Why not simply learn this arrangement?

To do this, we equip each individual NeRF $f_k$ with its own learnable affine transform $\mathbf{T}_k$, and denote the set of transforms across all volumes a layout $\mathbf{L} \equiv \{\mathbf{T}_k\}_{k=1}^K$. Each $\mathbf{T}_k$ has a rotation $\mathbf{R}_k \in \mathbb{R}^{3 \times 3}$ (in practice expressed via a quaternion $\mathbf{q} \in \mathbb{R}^4$ for ease of optimization), translation $\mathbf{t}_k \in \mathbb{R}^{3}$, and scale $s_k \in \mathbb{R}$. We apply this affine transform to the camera-to-world rays $\mathbf{r}$ before sampling the points used to query $f_k$. This implementation is simple, makes no assumptions about the underlying form of $f$, and updates parameters with standard backpropagation, as sampling and embedding points along the ray is fully differentiable \cite{lin2021barf}. Concretely, a ray $\mathbf{r}$ with origin $\mathbf{o}$ and direction $\mathbf{d}$ is transformed into an instance-specific ray $\mathbf{r}_k$ via the following transformations:
\begin{align}
    \mathbf{o}_k &= s_k \left(\mathbf{R}_k\mathbf{o} - \mathbf{t}_k\right) \\
    \mathbf{d}_k &= s_k \mathbf{R}_k \mathbf{d}\\
    \mathbf{r}_k(t) &= \mathbf{o}_k  + t \mathbf{d}_k  
\end{align}

\begin{figure*}[t]
    \input{figures/02_ablation}%
    \caption{\textbf{Evaluating disentanglement and quality.} We optimize a model with $K=3$ NeRFs on a list of 30 prompts, each containing three objects. We then automatically pair each NeRF with a description of one of the objects in the prompt and report average NeRF-object CLIP score (see text for details). We also generate each of the $30\times3=90$ objects from the prompt list individually and compute its score with both the corresponding prompt and a random other one, providing upper and lower bounds for performance on this task. Training $K$ NeRFs provides some decomposition, but most objects are scattered across 2 or 3 models. Learning one layout alleviates some of these issues, but only with multiple layouts do we see strong disentanglement. We show two representative examples of emergent objects to visualize these differences.}
    \label{fig:ablation}
     \vspace{-0.5em}
\end{figure*}

Though we input a different $H \times W$ grid of rays to each $f_k$, we composite their outputs as if they all sit in the same coordinate space---for example, the final density at $\boldsymbol{\mu} = \mathbf{r}(t)$ is the sum of densities output by every $f_k$ at $\boldsymbol{\mu}_k = \mathbf{r}_k(t)$.

Compared to the na\"ive formulation that instantiates $K$ models with identical initial densities, learning the size, orientation, and position of each model makes it easier to place density in different parts of 3D space. In addition, the inherent stochasticity of optimization may further dissuade degenerate solutions. %

While introducing layout learning significantly increases the quality of object disentanglement (Tbl. \ref{tbl:ablation_quant}), the model is still able to adjoin and utilize individual NeRFs in undesirable ways. For example, it can still place object parts next to each other in the same way as K NeRFs without layout learning. %

\begin{figure*}[t]
    \input{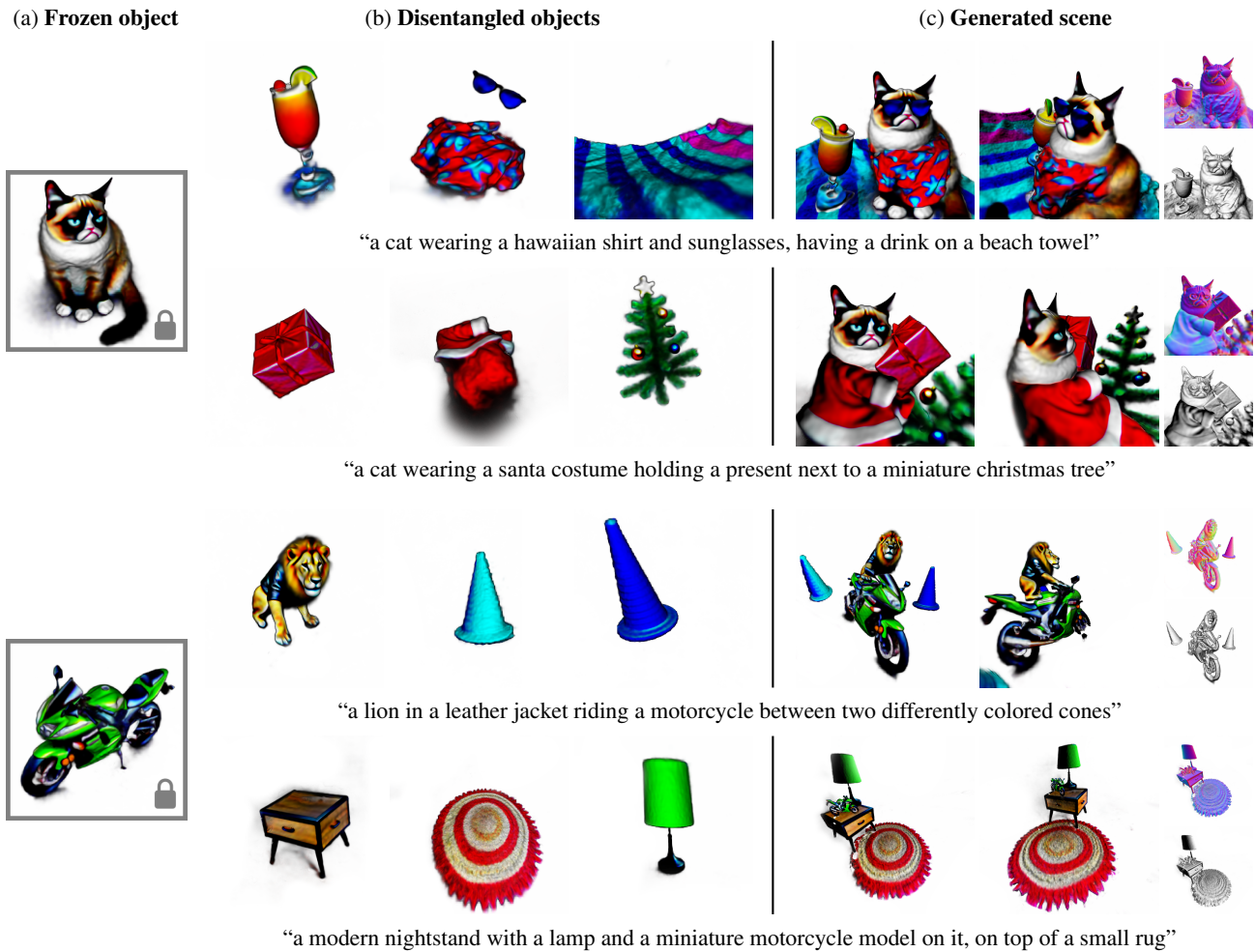}%
    \caption{\textbf{Conditional optimization.} We can take advantage of our structured representation to learn a scene given a 3D asset in addition to a text prompt, such as a specific cat or motorcycle \textbf{(a)}. By freezing the NeRF weights but not the layout weights, the model learns to arrange the provided asset in the context of the other objects it discovers \textbf{(b)}. We show the entire composite scenes the model creates in \textbf{(c)} from two views, along with surface normals and a textureless render.}
    \label{fig:partial_optim}
     \vspace{-0.75em}
\end{figure*}

\textbf{Learning multiple layouts.} We return to our statement that objects must be ``arranged in a particular way'' to form scenes that render to in-distribution images. While we already enable this with layout learning in its current form, we are not taking advantage of one key fact: there are many ``particular ways'' to arrange a set of objects, each of which gives an equally valid composition. Rather than only learning one layout, we instead learn a distribution over layouts $P(\mathbf{L})$ or a set of $N$ randomly initialized layouts $\{\mathbf{L}_n\}_{n=1}^N$. We opt for the latter, and sample one of the $N$ layouts from the set at each training step to yield transformed rays $\mathbf{r}_k$.

With this in place, we have arrived at our final definition of objectness (Figure \ref{fig:method}): \textbf{objects are parts of a scene that can be arranged in different ways to form valid compositions.} We have ``parts'' by incorporating multiple volumes, and ``arranging in different ways'' through multiple-layout learning. This simple approach is easy to implement (Fig. \ref{fig:pseudocode}), adds very few parameters ($8NK$ to be exact), requires no fine-tuning or manual annotation, and is agnostic to choices of text-to-image 
and 3D model. In Section~\ref{sec:experiments}, we verify that layout learning enables the generation and disentanglement of complex 3D scenes.%

\textbf{Regularization.} We build on Mip-NeRF 360 \cite{Barron_2022_CVPR} as our 3D backbone, inheriting their orientation, distortion, and accumulation losses to improve visual quality of renderings and minimize artifacts. However, rather than computing these losses on the final composited scene, we apply them on a per-NeRF basis. Importantly, we add a loss penalizing degenerate empty NeRFs by regularizing the soft-binarized version of each NeRF's accumulated density, ${\boldsymbol{\alpha}}_\text{bin}$, to occupy at least 10\% of the canvas:
\begin{align}
\label{eq:empty_loss}
    \mathcal{L}_\text{empty} &= \max\left(0.1 - \bar{\boldsymbol{\alpha}}_\text{bin}, 0\right)
\end{align}
We initialize parameters $s \sim \mathcal{N}(1, 0.3)$, $\mathbf{t}^{(i)} \sim \mathcal{N}(0, 0.3)$, and $\mathbf{q}^{(i)} \sim \mathcal{N}(\mu_{i}, 0.1)$ where $\mu_{i}$ is 1 for the last element and 0 for all others. We use a $10\times$ higher learning rate to train layout parameters. See Appendix~\ref{app:impl_dets} for more details.

\section{Experiments}
\label{sec:experiments}
We examine the ability of layout learning to generate and disentangle 3D scenes across a wide range of text prompts. We first verify our method's effectiveness through an ablation study and comparison to baselines, and then demonstrate various applications enabled by layout learning.

\vspace{-0.75em}
\subsection{Qualitative evaluation}
In Figure~\ref{fig:main_results}, we demonstrate several examples of our full system with layout learning. In each scene, we find that the composited 3D generation is high-quality and matches the text prompt, while the individual NeRFs learn to correspond to objects within the scene. Interestingly, since our approach does not directly rely on the input prompt, we can disentangle entities not mentioned in the text, such as a basket filled with easter eggs, a chef's hat, and a picnic table.
\vspace{-0.5em}

\subsection{Quantitative evaluation}
Measuring the quality of text-to-3D generation remains an open problem due to a lack of ground truth data---there is no ``true'' scene corresponding to a given prompt. Similarly, there is no true disentanglement for a certain text description. Following \citet{park2021benchmark,jain2022zeroshot,sds}, we attempt to capture both of these aspects using scores from a pretrained CLIP model \cite{radford2021learning, li2017learning}. Specifically, we create a diverse list of 30 prompts, each containing 3 objects,
and optimize a model with $K=3$ NeRFs on each prompt. We compute the 3$\times$3 matrix of CLIP scores ($100\times$ cosine similarity) for each NeRF with descriptions ``a DSLR photo of [object 1/2/3]'', finding the optimal NeRF-to-object matching and reporting the average score across all 3 objects. 

We also run SDS on the $30\times3=90$ per-object prompts individually and compute scores, representing a maximum attainable CLIP score under perfect disentanglement (we equalize parameter counts across all models for fairness). As a low-water mark, we compute scores between per-object NeRFs and a random other prompt from the pool of 90. 

The results in Table \ref{tbl:ablation_quant} show these CLIP scores, computed both on textured (``Color'') and textureless, geometry-only (``Geo'') renders. The final variant of layout learning achieves competitive performance, only 0.1 points away from supervised per-object rendering when using the largest CLIP model as an oracle, indicating high quality of both object disentanglement and appearance. Please see Appendix \ref{app:clip_eval} for a complete list of prompts and more details.

\textbf{Ablation.} We justify the sequence of design decisions presented in Section \ref{sec:method} by evaluating different variants of layout learning, starting from a simple collection of $K$ NeRFs and building up to our final architecture. The simple setting leads to some non-trivial separation (Figure \ref{fig:ablation_qual}) but parts of objects are randomly distributed across NeRFs---CLIP scores are significantly above random, but far below the upper bound. Adding regularization losses improve scores somewhat, but the biggest gains come from introducing layout learning and then co-learning $N$ different arrangements, validating our approach.

\subsection{Applications of layout learning}
To highlight the utility of the disentanglement given by layout learning beyond generation, we apply it to various 3D editing tasks. First, we show further results on object disentanglement in Figure \ref{fig:partial_optim}, but in a scenario where one NeRF is frozen to contain an object of interest, and the rest of the scene must be constructed around it. This object's layout parameters can also be frozen, for example, if a specific position or size is desired. We examine the more challenging setting where layout parameters must also be learned, and show results incorporating a grumpy cat and green motorbike into different contexts. Our model learns plausible transformations to incorporate provided assets into scenes, while still discovering the other objects necessary to complete the prompt.

\begin{figure}[t]
    \input{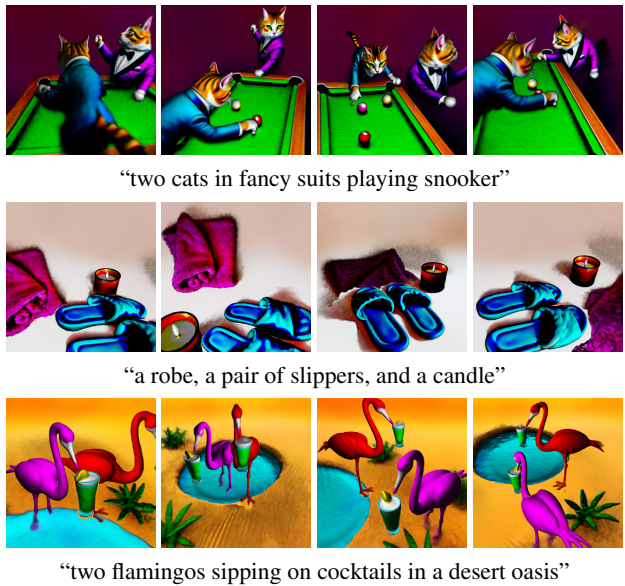}%
    \caption{\textbf{Layout diversity.} Our method discovers different plausible arrangements for objects. Here, we optimize each example over $N=4$ layouts and show differences in composited scenes, {\emph{e.g.}} flamingos wading inside vs. beside the pond, and cats in different poses around the snooker table.}
    \label{fig:diff_layouts}
     \vspace{-1.5em}
\end{figure}

In Figure \ref{fig:diff_layouts}, we visualize the different layouts learned in a single training run. %
The variation in discovered layouts is significant, indicating that our formulation can find various meaningful arrangements of objects in a scene. This allows users of our method to explore different permutations of the same content in the scenes they generate.

\begin{figure}[t]
    \input{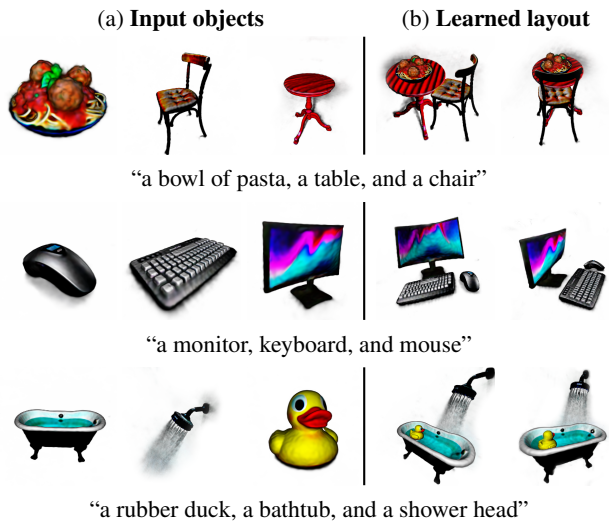}%
    \caption{\textbf{Optimizing layout.} Allowing gradients to flow only into layout parameters while freezing a set of provided 3D assets results in reasonable object configurations, such as a chair tucked into a table with spaghetti on it, despite no such guidance being provided in the text conditioning.}
    \label{fig:layout_learning}
     \vspace{-0.75em}
\end{figure}

Inspired by this, and to test gradient flow into layout parameters, we also examine whether our method can be used to arrange off-the-shelf, frozen 3D assets into semantically valid configurations (Figure \ref{fig:layout_learning}). Starting from random positions, sizes, and orientations, layouts are updated using signal backpropagated from the image model. This learns reasonable transformations, such as a rubber duck shrinking and moving inside a tub, and a shower head moving upwards and pointing so its stream is going into the tub.

\begin{figure*}[htpb!]
    \input{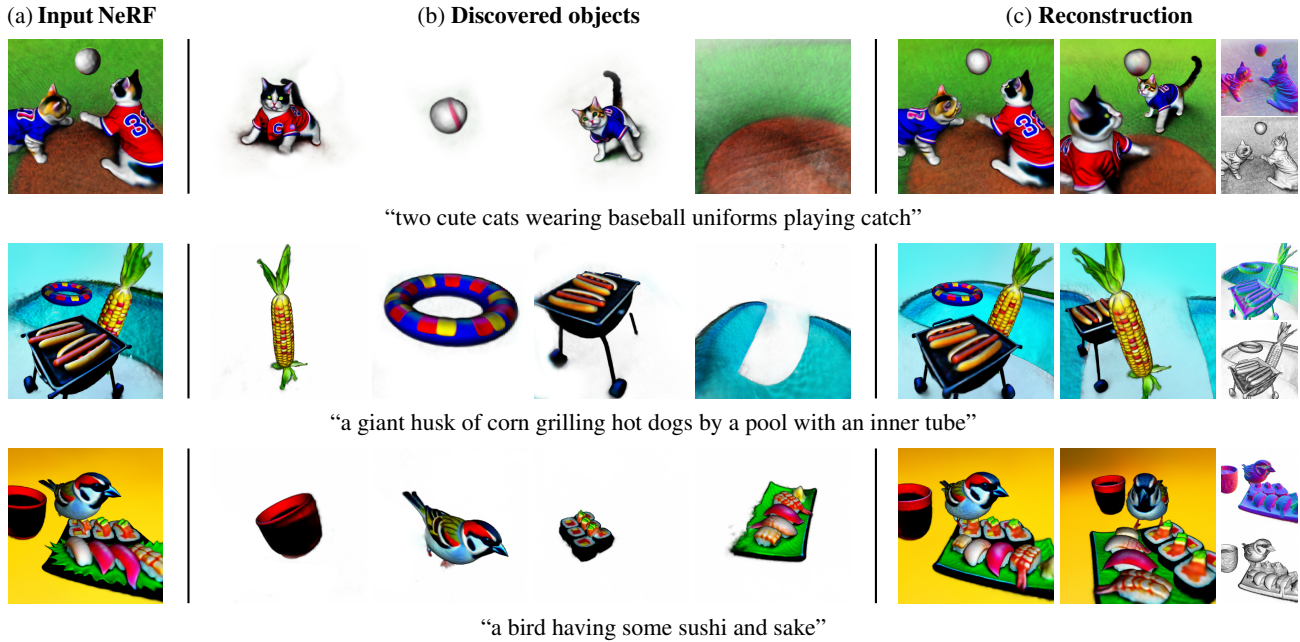}%
    \caption{\textbf{Decomposing NeRFs of scenes.} Given a NeRF representing a scene \textbf{(a)} and a caption, layout learning is able to parse the scene into the objects it contains without any per-object supervision \textbf{(b)}. We accomplish this by requiring renders of one of the $N$ learned layouts to match the same view rendered from the target NeRF \textbf{(c)}, using a simple $L_2$ reconstruction loss with $\lambda=0.05$.}
    \label{fig:decompositions}
\end{figure*}

Finally, we use layout learning to disentangle a pre-existing NeRF containing multiple entities, without any per-object supervision (Fig.~\ref{fig:decompositions}). We do this by randomly initializing a new model and training it with a caption describing the target NeRF. We require the first layout $\mathbf{L}_1$ to create a scene that faithfully reconstructs the target NeRF in RGB space, allowing all other layouts to vary freely. We find that layout learning arrives at reasonable decompositions of the scenes it is tasked with reconstructing.

\begin{figure}[t]
\captionsetup{singlelinecheck=false}
    \input{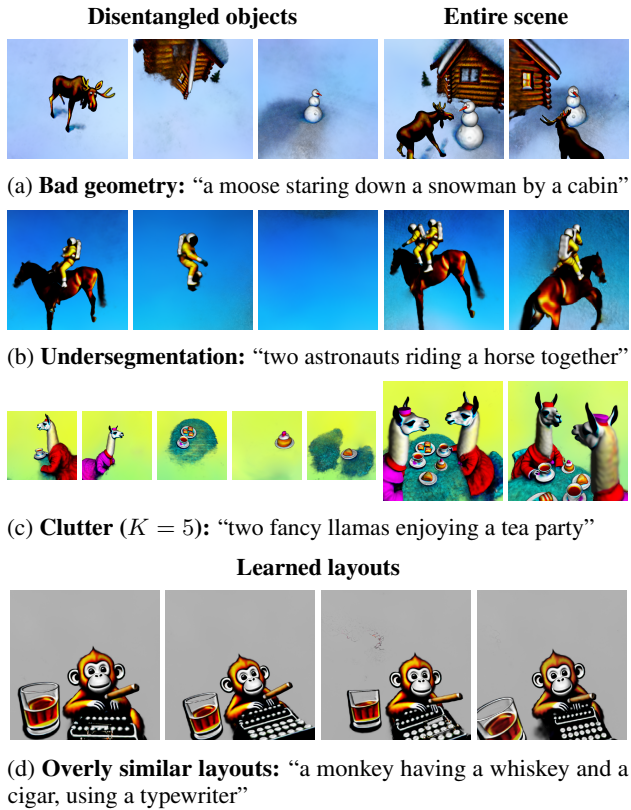}%
    \caption{\textbf{Limitations.} Layout learning inherits failure modes from SDS, such as bad geometry of a cabin with oddly intersecting exterior walls \textbf{(a)}. It also may undesirably group objects that always move together \textbf{(b)} such as a horse and its rider, and \textbf{(c)} for certain prompts that generate many small objects, choosing $K$ correctly is challenging, hurting disentanglement. In some cases \textbf{(d)}, despite different initial values, layouts converge to very similar final configurations.
    }
    \label{fig:limitations}
     \vspace{-1em}
\end{figure}
\section{Related work}
\textbf{Object recognition and discovery.} The predominant way to identify the objects present in a scene is to segment two-dimensional images using extensive manual annotation \cite{kirillov2023segment, li2022exploring, wang2023yolov7}, but relying on human supervision introduces challenges and scales poorly to 3D data. As an alternative, an extensive line of work on   {\em unsupervised} object discovery \cite{russell2006using,rubinstein2013unsupervised,oktay2018counterfactual,
henaff2022object,smith2022unsupervised,ye2022deformable,dbw} proposes different inductive biases \cite{locatello2019challenging} that encourage awareness of objects in a scene. However, these approaches are largely restricted to either 2D images or constrained 3D data \cite{yu2021unsupervised,sajjadi2022object}, limiting their applicability to complex 3D scenes. 
At the same time, large text-to-image models have been shown to implicitly encode an understanding of entities in their internals \cite{epstein2023selfguidance}, motivating their use for the difficult problem of explicit object disentanglement. 

\textbf{Compositional 3D generation.} There are many benefits to generating 3D scenes separated into objects beyond just better control. For example, generating objects one at a time and compositing them manually provides no guarantees about compatibility in appearance or pose, such as ``dogs in matching outfits'' in Figure~\ref{fig:main_results} or a lion holding the handlebars of a motorcycle in Figure~\ref{fig:partial_optim}.  
 Previous and concurrent work explores this area, but either requires users to painstakingly annotate 3D bounding boxes and per-object labels \cite{cohenbar2023setthescene, po2023compositional} or uses external supervision such as LLMs to propose objects and layouts \cite{yang2023holodeck, zhang2023scenewiz3d}, significantly slowing down the generation process and hindering quality. We show that this entire process can be solved without any additional models or labels, simply using the signal provided by a pretrained image generator. 
\section{Discussion}
We present layout learning, a simple method for generating disentangled 3D scenes given a text prompt. By optimizing multiple NeRFs to form valid scenes across multiple layouts, we encourage each NeRF to contain its own object. This approach requires no additional supervision or auxiliary models, yet performs quite well. By generating scenes that are decomposed into objects, we provide users of text-to-3D systems with more granular, local control over the complex creations output by a black-box neural network. 

Though layout learning is surprisingly effective on a wide variety of text prompts, the problem of object disentanglement in 3D is inherently ill-posed, and our definition of objects is simple. As a result, many undesirable solutions exist that satisfy the constraints we pose. 

Despite our best efforts, the compositional scenes output by our model do occasionally suffer from failures (Fig.~\ref{fig:limitations}) such as over- or under-segmentation and the ``Janus problem'' (where objects are depicted so that salient features appear from all views, {\em e.g.} an animal with a face on the back of its head) as well as other undesirable geometries. Further, though layouts are initialized with high standard deviation and trained with an increased learning rate, they occasionally converge to near-identical values, minimizing the effectivness of our method. In general, we find that failures to disentangle are accompanied by an overall decrease in visual quality.

\section*{Acknowledgements}
We thank Dor Verbin, Ruiqi Gao, Lucy Chai, and Minyoung Huh for their helpful comments, and Arthur Brussee for help with an NGP implementation. DE was
            partly supported by the PD Soros Fellowship. DE conducted part of this research at Google, with additional
            funding from an ONR MURI grant.
\section*{Impact statement}

Generative models present many ethical concerns over data attribution, nefarious applications, and longer-term societal effects. Though we build on a text-to-image model trained on data that has been filtered to remove concerning imagery and captions, recent work has shown that popular datasets contain dangerous depictions of undesirable content\footnote{https://crsreports.congress.gov/product/pdf/R/R47569} which may leak into model weights. 

Further, since we distill the distribution learned by an image generator, we inherit the potential negative use-cases enabled by the original model. By facilitating the creation of more complex, compositional 3D scenes, we perhaps expand the scope of potential issues associated with text-to-3D technologies. Taking care to minimize potential harmful deployment of our generative models through using ethically-sourced and well-curated data is of the utmost importance as our field continues to grow in size and influence.

Further, by introducing an unsupervised method to disentangle 3D scenes into objects, we possibly contribute to the displacement of creative workers such as video game asset designers via increased automation. However, at the same time, methods like the one we propose have the potential to become valuable tools at the artist's disposal, providing much more control over outputs and helping create new, more engaging forms of content.

\bibliography{example_paper}
\bibliographystyle{icml2024}

\FloatBarrier
\appendix
\section{Appendix}

\subsection{Implementation details}
\label{app:impl_dets}
We use Mip-NeRF 360 as the 3D backbone \cite{Barron_2022_CVPR} and Imagen \cite{saharia2022photorealistic}, a 128px pixel-space diffusion model, for most experiments, rendering at 512px. To composite multiple representations, we merge the output albedos and densities at each point, taking the final albedo as a weighted average given by per-NeRF density. We apply this operation to the outputs of the proposal MLPs as well as the final RGB-outputting NeRFs. We use $\lambda_\text{dist}=0.001, \lambda_\text{acc}=0.01, \lambda_\text{ori}=0.01$ as well as $\lambda_\text{empty} = 0.05$. The empty loss examines the mean of the per-pixel accumulated density along rays in a rendered view, $\boldsymbol{\alpha}$, for each NeRF. It penalizes these mean $\bar{\boldsymbol{\alpha}}$ values if they are under a certain fraction of the image canvas (we use 10\%). For more robustness to noise, we pass $\boldsymbol{\alpha}$ through a scaled sigmoid to binarize it (Fig.~\ref{fig:soft_bin_acc_pseudocode}), yielding the $\bar{\boldsymbol{\alpha}}_\text{bin}$ used in Eq.~\ref{eq:empty_loss}.  We sample camera azimuth in $[0^\circ,360^\circ]$ and elevation in $[-90^\circ,0^\circ]$ except in rare cases where we sample azimuth in a 90-degree range to minimize Janus-problem artifacts or generate indoor scenes with a diorama-like effect.

We use a classifier-free guidance strength of 200 and textureless shading probability of $0.1$ for SDS \cite{sds}, disabling view-dependent prompting as it does not aid in the generation of compositional scenes (Table~\ref{tbl:ablation_quant}). We otherwise inherit all other details, such as covariance annealing and random background rendering, from SDS. We optimize our model with Shampoo \cite{gupta2018shampoo} with a batch size of 1 for 15000 steps with an annealed learning rate, starting from $10^{-9}$, peaking at $10^{-4}$ after 3000 steps, and decaying to $10^{-6}$. 

\textbf{Optimizing NGPs.} To verify the robustness of our approach to different underlying 3D representations, we also experiment with a re-implementation of Instant NGPs \cite{mueller2022instant}, and find that our method generalizes to that setting. Importantly, we implement an aggressive coarse-to-fine training regime in the form of slowly unlocking grid settings at resolution higher than $64\times64$ only after 2000 steps. Without this constraint on the initial smoothness of geometry, the representation ``optimizes too fast'' and is prone to placing all density in one NGP.

\subsection{Pseudo-code for layout learning}
\label{app:pseudocode}

In Figs. \ref{fig:pseudocode} and \ref{fig:soft_bin_acc_pseudocode}, we provide NumPy-like pseudocode snippets of the core logic necessary to implement layout learning, from transforming camera rays to compositing multiple 3D volumes to regularizing them.
\begin{figure}[t!]
\begin{lrbox}{\mintedbox}
\RecustomVerbatimEnvironment{Verbatim}{BVerbatim}{}
    \begin{minted}[
frame=leftline,
framesep=2mm,
fontsize=\footnotesize,
]{python}
# Initialize variables
quat = normal((N, K, 4), mean=[0,0,0,1.], std=0.1)
trans = normal((N, K, 3), mean=0., std=0.3)
scale = normal((N, K, 1), mean=1., std=0.3)
nerfs = [init_nerf() for i in range(K)]

# Transform rays for NeRF k using layout n
def transform(rays, k, n):
  rot = quaternion_to_matrix(quat)
  rays['orig'] = rot[n,k] @ rays['orig'] - trans[n,k]
  rays['orig'] *= scale[n,k]
  rays['dir'] = scale[n,k] * rot[n,k] @ rays['dir']
  return rays

# Composite K NeRFs into one volume
def composite_nerfs(per_nerf_rays):
  per_nerf_out = [nerf(rays) for nerf, rays 
    in zip(nerfs, per_nerf_rays]
  densities = [out['density'] for out in per_nerf_out]
  out = {'density': sum(densities)}
  wts = [d/sum(densities) for d in densities]
  rgbs = [out['rgb'] for out in per_nerf_out]
  out['rgb'] = sum(w*rgb for w,rgb in zip(wts, rgbs))
  return out, per_nerf_out

# Train
optim = shampoo(params=[nerfs, quat, trans, scale])
for step in range(num_steps):
  rays = sample_camera_rays()
  n = random.uniform(N)
  per_nerf_rays = [
    transform(rays, k, n) for k in range(K)
  ]
  vol, per_nerf_vols = composite_nerfs(per_nerf_rays)
  image = render(vol, rays)
  loss = SDS(image, prompt, diffusion_model)
  loss += regularize(per_nerf_vols)
  loss.backward()
  optim.step_and_zero_grad()
\end{minted}
\end{lrbox}

\resizebox{\columnwidth}{!}{\usebox{\mintedbox}}
    \caption{\textbf{Pseudocode for layout learning}, with segments inherited from previous work abstracted into functions.}
    \label{fig:pseudocode}
\end{figure}

\subsection{CLIP evaluation}
\label{app:clip_eval}
\begin{figure}[t]
    \begin{minted}[
fontsize=\footnotesize,
]{python}
def soft_bin(x, t=0.01, eps=1e-7):
  # x has shape (..., H, W)
  bin = sigmoid((x - 0.5)/t)
  min = bin.min(axis=(-1, -2), keepdims=True)
  max = bin.max(axis=(-1, -2), keepdims=True)
  return (bin - min) / (max - min + eps)
soft_bin_acc = soft_bin(acc).mean((-1,-2))
empty_loss = empty_loss_margin - soft_bin_acc
empty_loss = max(empty_loss, 0.)
\end{minted}
     \vspace{-1.em}

    \caption{\textbf{Pseudocode for empty NeRF regularization,} where \texttt{soft\_bin\_acc} computes $\bar{\boldsymbol{\alpha}}_\text{bin}$ in Equation \ref{eq:empty_loss}.}
    \label{fig:soft_bin_acc_pseudocode}
\end{figure}

\begin{figure}[h]
\begin{lrbox}{\mintedbox}
\RecustomVerbatimEnvironment{Verbatim}{BVerbatim}{}
    \begin{minted}[
frame=leftline,
framesep=2mm,
fontsize=\footnotesize,
]{python}
'a cup of coffee, a croissant, and a closed book',
'a pair of slippers, a robe, and a candle',
'a basket of berries, a carton of whipped cream, and an orange',
'a guitar, a drum set, and an amp',
'a campfire, a bag of marshmallows, and a warm blanket',
'a pencil, an eraser, and a protractor',
'a fork, a knife, and a spoon',
'a baseball, a baseball bat, and a baseball glove',
'a paintbrush, an empty easel, and a palette',
'a teapot, a teacup, and a cucumber sandwich',
'a wallet, keys, and a smartphone',
'a backpack, a water bottle, and a bag of chips',
'a diamond, a ruby, and an emerald',
'a pool table, a dartboard, and a stool',
'a tennis racket, a tennis ball, and a net',
'sunglasses, sunscreen, and a beach towel',
'a ball of yarn, a pillow, and a fluffy cat',
'an old-fashioned typewriter, a cigar, and a glass of whiskey',
'a shovel, a pail, and a sandcastle',
'a microscope, a flask, and a laptop',
'a sunny side up egg, a piece of toast, and some strips of bacon',
'a vase of roses, a slice of chocolate cake, and a bottle of red wine',
'three playing cards, a stack of poker chips, and a flute of champagne',
'a tomato, a stalk of celery, and an onion',
'a coffee machine, a jar of milk, and a pile of coffee beans',
'a bag of flour, a bowl of eggs, and a stick of butter',
'a hot dog, a bottle of soda, and a picnic table',
'a pothos houseplant, an armchair, and a floor lamp',
'an alarm clock, a banana, and a calendar',
'a wrench, a hammer, and a measuring tape',
'a backpack, a bicycle helmet, and a watermelon'
\end{minted}
\end{lrbox}
\resizebox{\columnwidth}{!}{\usebox{\mintedbox}}
    \caption{\textbf{Prompts used for CLIP evaluation.} Each prompt is injected into the template ``a DSLR photo of \{prompt\}, plain solid color background''. To generate individual objects, the three objects in each prompt are separated into three new prompts and optimized independently.}
    \label{fig:clip_prompts}
\end{figure}

To evaluate our approach, we use similarity scores output by a pretrained contrastive text-image model \cite{radford2021learning}, which have been shown to correlate with human judgments on the quality of compositional generation \cite{park2021benchmark}. However, rather than compute a retrieval-based metric such as precision or recall, we report the raw ($100\times$ upscaled, as is common practice) cosine similarities. In addition to being a more granular metric, this avoids the dependency of retrieval on the size and difficulty of the test set (typically only a few hundred text prompts).

We devise a list of 30 prompts (Fig.~\ref{fig:clip_prompts}), each of which lists three objects, spanning a wide range of data, from animals to food to sports equipment to musical instruments. As described in Section~\ref{sec:experiments}, we then train models with $K=3$ NeRFs and layout learning and test whether each NeRF contains a different object mentioned in the prompt. We compute CLIP scores for each NeRF with a query prompt  ``a DSLR photo of [A/B/C]'', yielding a $3\times3$ score matrix.

To compute NeRF-prompt CLIP scores, we average text-image similarity across 12 uniformly sampled views, each 30 degrees apart, at $-30^\circ$ elevation. We then select the best NeRF-prompt assignment (using brute force, as there are only $3! = 6$ possible choices), and run this process across 3 different seeds, choosing the one with the highest mean NeRF-prompt score.

\end{document}